\documentclass[sigconf]{acmart}

\usepackage{multirow}
\usepackage{booktabs}
\usepackage{enumitem}
\AtBeginDocument{%
  }

\setcopyright{acmlicensed}
\copyrightyear{2027}
\acmYear{2027}
\acmDOI{XXXXXXX.XXXXXXX}
\acmConference[KDD '27]{ACM SIGKDD Conference on Knowledge Discovery and Data Mining}{August 1--5, 2027}{San Jose, CA, USA}
\begin{document}
\title{TaxDistill: Improving Metagenomic Taxonomic Annotation via Distilled Genomic Foundation Models}

\author{Rongye Ye}
\email{yerongye@big.ac.cn}
\authornote{These authors contributed equally to this work.}
\affiliation{%
  \institution{National Genomics Data Center, China National Center for Bioinformation}
  \city{Beijing}
  \postcode{100101}
  \country{China}}
\affiliation{%
  \institution{Beijing Institute of Genomics, Chinese Academy of Sciences}
  \city{Beijing}
  \postcode{100101}
  \country{China}}
  
\affiliation{%
  \institution{University of Chinese Academy of Sciences}
  \city{Beijing}
  \postcode{100049}
  \country{China}}

\author{Lun Li}
\email{lil@big.ac.cn}
\authornotemark[1]
\affiliation{%
  \institution{National Genomics Data Center, China National Center for Bioinformation}
  \city{Beijing}
  \postcode{100101}
  \country{China}}
\affiliation{%
  \institution{Beijing Key Laboratory of Intelligent Governance and Application of Biological Big Data, China National Center for Bioinformation}
  \city{Beijing}
  \postcode{100049}
  \country{China}}
\affiliation{%
  \institution{Beijing Institute of Genomics, Chinese Academy of Sciences}
  \city{Beijing}
  \postcode{100101}
  \country{China}}

\author{Zheng Luo}
\email{luozheng@big.ac.cn}
\affiliation{%
  \institution{National Genomics Data Center, China National Center for Bioinformation}
  \city{Beijing}
  \postcode{100101}
  \country{China}}
\affiliation{%
  \institution{Beijing Institute of Genomics, Chinese Academy of Sciences}
  \city{Beijing}
  \postcode{100101}
  \country{China}}
\affiliation{%
  \institution{University of Chinese Academy of Sciences}
  \city{Beijing}
  \postcode{100049}
  \country{China}}

\author{Yiran Zhan}
\email{zhanyiran@big.ac.cn}
\affiliation{%
  \institution{National Genomics Data Center, China National Center for Bioinformation}
  \city{Beijing}
  \postcode{100101}
  \country{China}}
\affiliation{%
  \institution{Beijing Institute of Genomics, Chinese Academy of Sciences}
  \city{Beijing}
  \postcode{100101}
  \country{China}}
\affiliation{%
  \institution{University of Chinese Academy of Sciences}
  \city{Beijing}
  \postcode{100049}
  \country{China}}

\author{Zhang Zhang}
\email{zhangzhang@big.ac.cn}
\affiliation{%
  \institution{National Genomics Data Center, China National Center for Bioinformation}
  \city{Beijing}
  \postcode{100101}
  \country{China}}
\affiliation{%
  \institution{Beijing Key Laboratory of Intelligent Governance and Application of Biological Big Data, China National Center for Bioinformation}
  \city{Beijing}
  \postcode{100049}
  \country{China}}
\affiliation{%
  \institution{Beijing Institute of Genomics, Chinese Academy of Sciences}
  \city{Beijing}
  \postcode{100101}
  \country{China}}
\affiliation{%
  \institution{University of Chinese Academy of Sciences}
  \city{Beijing}
  \postcode{100049}
  \country{China}}

\author{Shuhui Song}
\authornote{Corresponding author.}
\email{songshh@big.ac.cn}
\affiliation{%
  \institution{National Genomics Data Center, China National Center for Bioinformation}
  \city{Beijing}
  \postcode{100101}
  \country{China}}
\affiliation{%
  \institution{Beijing Key Laboratory of Intelligent Governance and Application of Biological Big Data, China National Center for Bioinformation}
  \city{Beijing}
  \postcode{100049}
  \country{China}}
\affiliation{%
  \institution{Beijing Institute of Genomics, Chinese Academy of Sciences}
  \city{Beijing}
  \postcode{100101}
  \country{China}}
\affiliation{%
  \institution{University of Chinese Academy of Sciences}
  \city{Beijing}
  \postcode{100049}
  \country{China}}


\renewcommand{\shortauthors}{Ye and Li et al.}

\begin{abstract}
Metagenomic taxonomic annotation is essential for interpreting complex microbial communities, yet reliable annotation remains challenging under reference database incompleteness and ambiguous taxonomic boundaries. Existing similarity-based tools are efficient, but they often produce noisy pseudo-labels in complex environments; learning-based post-hoc correction methods can further inherit this noise when trained on hard pseudo-labels generated by upstream classifiers. We propose TaxDistill, a plug-and-play knowledge distillation framework for reliable metagenomic taxonomic recalibration. TaxDistill uses the genomic foundation model GenomeOcean as a semantic teacher and performs one-way online distillation, where the teacher is updated by hierarchical taxonomic supervision while its detached soft probability distributions guide a lightweight student network. This design mitigates the student's overfitting to noisy hard pseudo-labels and supports confidence-based rejection of potentially unreliable predictions. Comprehensive experiments on seven diverse CAMI2 datasets with MMseqs2, Metabuli, and Kraken2 show that TaxDistill improves annotation reliability over existing baselines in most scenarios. For example, on the Gastrointestinal dataset, TaxDistill improves the F1 score of MMseqs2 from 0.763 to 0.941, surpassing the Taxometer baseline. Our results suggest that large genomic foundation models can provide semantic supervision for noisy scientific annotations, enabling lightweight post-hoc recalibration of existing metagenomic annotation results.
\end{abstract}





\maketitle

\section{Introduction}

Metagenomic sequencing has become a fundamental technology for profiling complex microbial communities, enabling researchers to decipher the "language of life" embedded in environmental, clinical, and host-associated samples \citep{handelsman2004metagenomics,prabakaran2025deciphering,levy2025cutting}. A central step in this process is taxonomic annotation, which maps sequencing reads or assembled contigs to taxonomic nodes and thereby supports downstream tasks such as clinical pathogen detection and disease-associated microbiome characterization \citep{simon2019benchmarking,chiu2019clinical}. In practice, reliable taxonomic annotation remains challenging because real metagenomic samples often contain rare, novel, or uncultured organisms that are poorly represented in reference databases \citep{meyer2022critical}. As a result, the scientific challenge is not merely to assign every sequence to a known taxon, but to produce reliable annotations under reference incompleteness, ambiguous taxonomic boundaries, and substantial label noise.

\begin{figure}[t]
    \centering
    \includegraphics[width=\columnwidth]{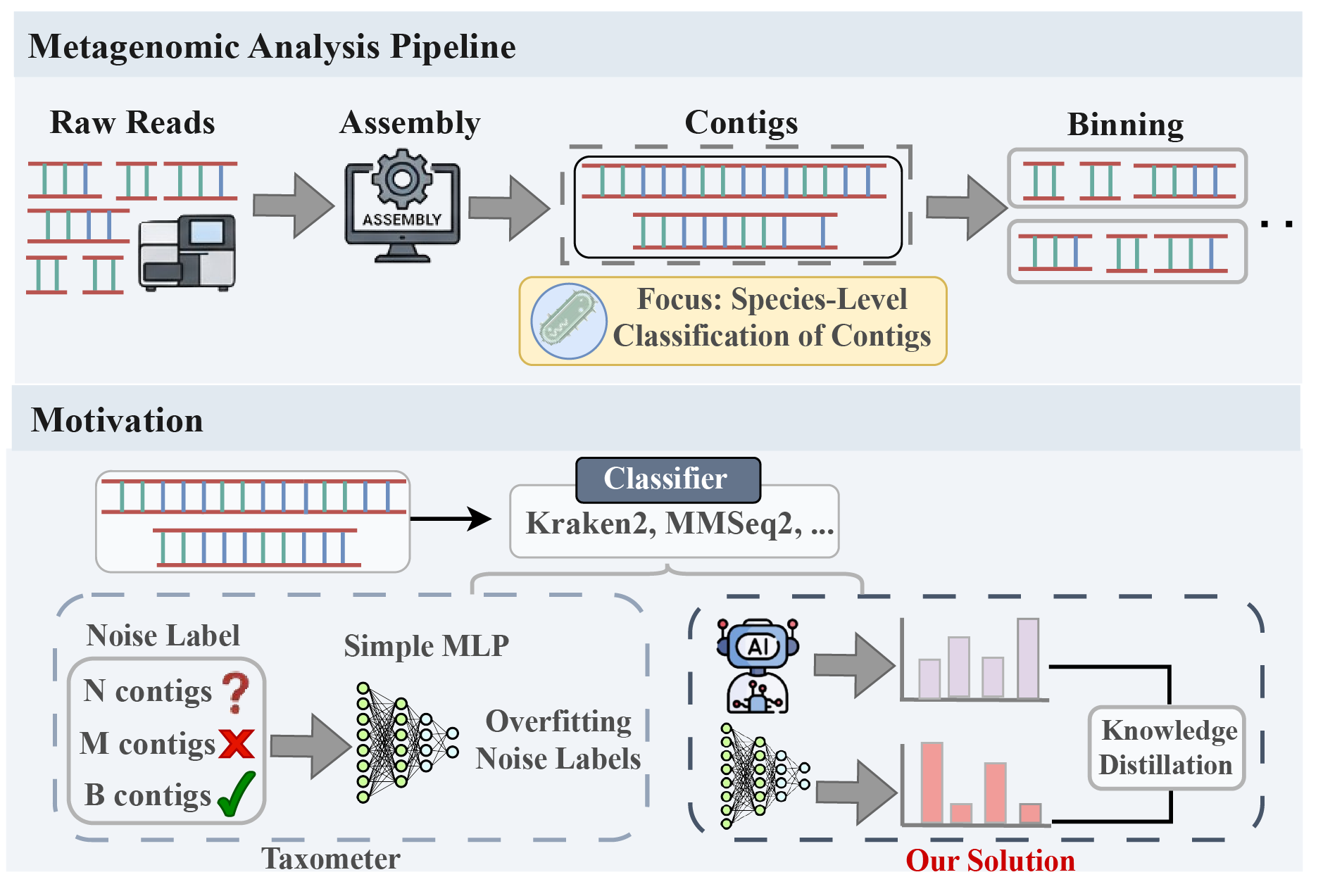}
    \caption{Metagenomic Analysis Pipeline and Research Motivation}
    \label{Figure1}
\end{figure}

Current mainstream annotation tools primarily rely on sequence similarity search or heuristic matching, including Kraken2 \citep{wood2019improved}, MMseqs2 \citep{kallenborn2025gpu}, and Metabuli \citep{kim2024metabuli}. These tools are efficient and effective for well-characterized microbes, but often produce erroneous assignments or unassigned contigs in complex microbial environments. Recent advances in deep learning for sequence modeling \citep{vaswani2017attention,ye2025fly} have motivated learning-based approaches that attempt to recalibrate such predictions. A representative example is Taxometer \citep{kutuzova2024taxometer}, which combines tetranucleotide frequencies (TNFs) and abundance information and uses a deep hierarchical loss \citep{valmadre2022hierarchical} to align predictions with the taxonomic tree. Taxometer establishes an important post-hoc correction paradigm, but it still depends on noisy pseudo-labels generated by initial retrieval tools. Since its feature encoder is a lightweight Multilayer Perceptron (MLP), the model can overfit erroneous hard labels and suffer from severe noise memorization \citep{zhang2016understanding,liu2020early,vishwakarma2025rethinking}, limiting its ability to perform robust self-correction.

Genomic language models provide a new opportunity to address this reliability bottleneck. These models have demonstrated broad potential for applications in the life sciences \citep{lin2023evolutionary,brixi2026genome,cheng2024training,ye2026influ}. Among them, GenomeOcean is a generative language model pre-trained on more than 600 Gbp of large-scale, complex metagenomic assembled sequences. Similar to large language models in NLP, GenomeOcean \citep{zhou2025genomeocean} adopts an efficient Byte Pair Encoding (BPE) tokenization strategy to construct its genomic vocabulary, allowing it to capture implicit grammatical constraints in DNA sequences and model complex long-range dependencies. However, directly deploying such a foundation model as a taxonomic classifier is not always practical or optimal, due to input length constraints, computational overhead, and the fact that fine-tuning only the classification head often fails to capture dataset-specific taxonomic structures, leading to suboptimal adaptation. We therefore ask a different question: can a genomic foundation model serve as a semantic regularizer that improves the reliability of lightweight post-hoc taxonomic recalibration?

Driven by the research objectives illustrated in Figure \ref{Figure1}, we introduce TaxDistill, a knowledge distillation framework for reliable contig-level metagenomic taxonomic annotation. TaxDistill does not aim to replace existing retrieval-based classifiers. Instead, it serves as a plug-and-play post-hoc label-denoising tool that recalibrates their noisy predictions. Specifically, we retain the efficient lightweight architecture used in Taxometer as the student network and introduce GenomeOcean as the teacher model. Rather than treating the teacher as an infallible classifier, TaxDistill distills its soft probability distributions as instance-specific regularization. We formulate this process as one-way online distillation: the teacher branch is updated during training by the hierarchical classification objective, while the distillation loss supervises only the student branch through detached teacher outputs. This design transfers relational genomic knowledge into the student model while reducing overfitting to noisy hard pseudo-labels. Experiments on diverse CAMI2 datasets show that TaxDistill consistently improves annotation reliability in complex microbial environments, especially by rescuing unassigned contigs and reducing false-positive assignments through conservative confidence-based recalibration In summary, our main contributions are as follows:

\begin{enumerate}[label=\arabic*., leftmargin=*]
    \item We propose TaxDistill, a plug-and-play post-hoc recalibration framework for metagenomic taxonomic annotation under noisy pseudo-labels. It can be integrated with different retrieval-based classifiers without changing their original inference pipelines.
    
    \item We introduce genomic foundation model distillation into metagenomic label correction. By using GenomeOcean as a semantic teacher rather than an oracle classifier, TaxDistill performs one-way online distillation that transfers softened taxonomic supervision and relational genomic knowledge to a lightweight student. This design mitigates overfitting to noisy hard labels and supports confidence-based recalibration that assigns potentially unreliable predictions to the unclassified category for improved false-positive control.

    \item We conduct comprehensive benchmarking on seven diverse CAMI2 microbial environment datasets, evaluating multiple mainstream classifiers (MMseqs2 \citep{kallenborn2025gpu}, Metabuli \citep{kim2024metabuli}, Kraken2 \citep{wood2019improved}) and the existing calibration model Taxometer \citep{kutuzova2024taxometer}. Experimental results demonstrate that TaxDistill outperforms baseline models in most scenarios.
\end{enumerate}

\section{Related Work}

\subsection{Metagenomic Taxonomic Annotation}

Metagenomic taxonomic annotation has traditionally been dominated by sequence similarity search and heuristic matching. Representative tools include Kraken2 \citep{wood2019improved}, MMseqs2 \citep{kallenborn2025gpu}, and the recently proposed Metabuli \citep{kim2024metabuli}. These methods are computationally efficient and effective for well-characterized organisms, but they remain sensitive to reference incompleteness. In complex environmental samples or for novel microorganisms absent from reference databases, they can produce erroneous labels, ambiguous assignments, or unassigned sequences.

\subsection{Learning-Based Methods}
Learning-based methods have been introduced to improve metagenomic sequence representation beyond direct sequence matching. DeepMicrobes \citep{liang2020deepmicrobes} uses bidirectional long short-term memory (Bi-LSTM) units, while MetaTransformer \citep{wichmann2023metatransformer} leverages self-attention mechanisms. However, many end-to-end classifiers rely on simulated short reads and fixed reference labels, making them vulnerable to closed-set assumptions because the data distribution encountered in real metagenomic sequencing often differs substantially from the training distribution \citep{nayfach2021genomic,thompson2017communal}. Taxometer \citep{kutuzova2024taxometer} addresses part of this limitation by constructing a local label tree from the output of an initial classifier and performing post-hoc label correction. Nevertheless, because it relies on lightweight handcrafted features and pseudo-labels from retrieval tools, it remains susceptible to overfitting noisy initial annotations.

\subsection{Knowledge Distillation Methods}

Knowledge distillation (KD) transfers the representation capability of a complex teacher model to a compact student model \citep{hinton2015distilling,gou2021knowledge}. Prior studies have shown that KD can alleviate learning from noisy labels because soft teacher distributions encode inter-class relationships and act as a regularizer, unlike hard labels that force overconfident supervision \citep{li2017learning,muller2019does,yuan2020revisiting,ben2024distilling}. Meanwhile, genomic foundation models are emerging in computational biology and metagenomics, offering rich sequence representations learned from large-scale genomic corpora. However, how to distill such foundation-model knowledge into lightweight networks for noisy metagenomic taxonomic recalibration remains underexplored. TaxDistill is designed to address this gap.

\section{Methodology}
\label{sec:methodology}

\begin{figure*}[t]
    \centering
    \includegraphics[width=\textwidth]{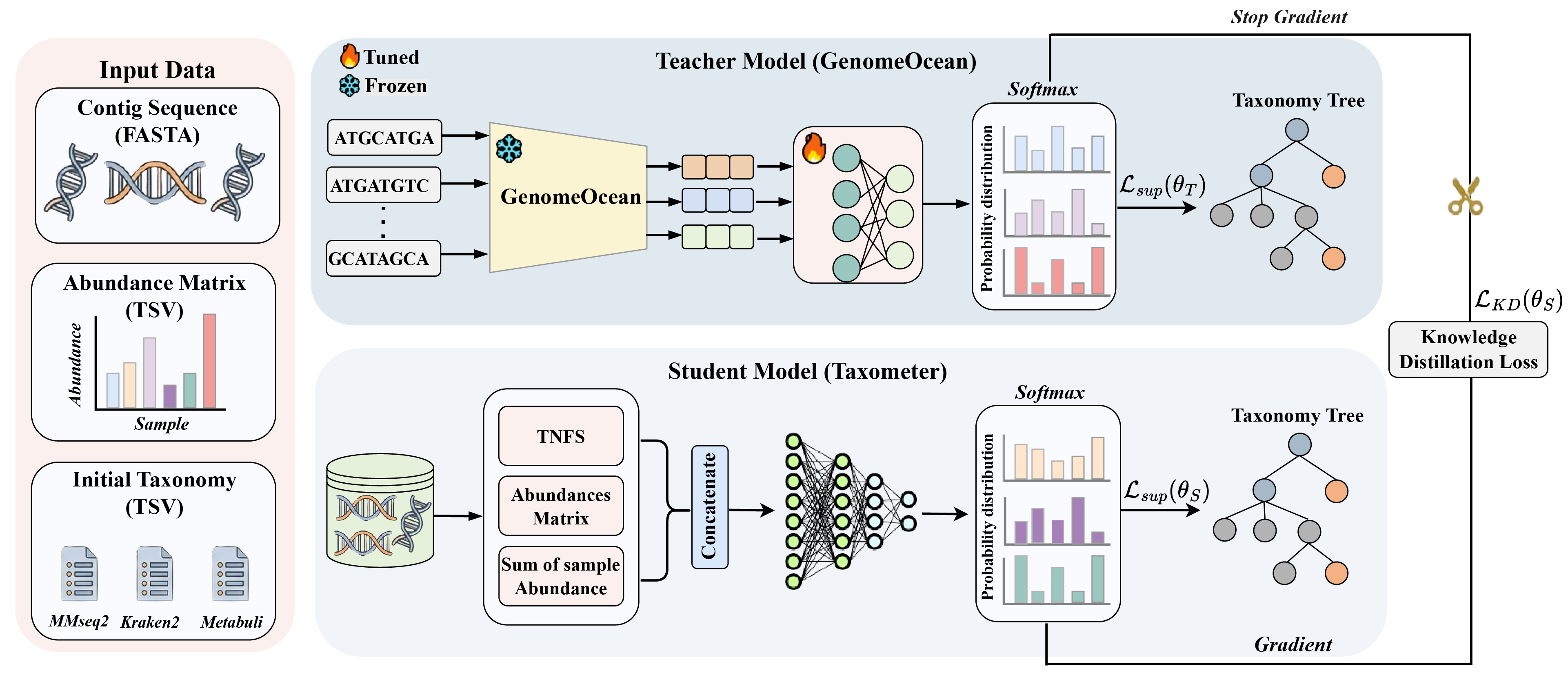}
    \caption{The overall architecture of the proposed TaxDistill framework. It consists of three core modules: multimodal data input formulation, the teacher branch, and the student branch.}
    \label{figure2}
\end{figure*}

In this section, we formally introduce the proposed TaxDistill framework. As illustrated in Figure \ref{figure2}, the teacher branch employs the pre-trained GenomeOcean foundation model with a frozen backbone. It extracts deep semantic features from raw sequences and projects them through a learnable classification head to output a categorical probability distribution. This branch is optimized independently via a deep hierarchical loss.

The student branch maintains a lightweight MLP architecture to ensure inference with low latency. It processes a $(103 + K + 1)$ dimensional feature vector consisting of hand-crafted TNFs-based features, abundances across K environments, and total abundance.

During the joint optimization phase, the KD loss is introduced to quantify the divergence between the teacher and student distributions. The student model's parameters are jointly updated by its own hierarchical classification loss and the KD loss. Meanwhile, the teacher model continues to be updated solely by its classification loss. The detailed mathematical formulations are presented below.

\subsection{Problem Formulation and Notation}
\label{sec:problem_formulation}

We first define a directed hierarchical taxonomic tree  $\mathcal{T} = (\mathcal{V}, \mathcal{E})$ for each target dataset, where $\mathcal{V}$ is the set of all taxonomic nodes, and $\mathcal{E}$ is the set of directed edges representing the parent-child hierarchical relationships (i.e., $(u, v) \in \mathcal{E}$ indicates that $u$ is the direct parent of $v$). Let $\mathcal{N} \subset \mathcal{V}$ denote the set of leaf nodes representing fine-grained taxonomic labels. For any node $u \in \mathcal{V}$, we define $leaves(u)$ as the set of all leaf nodes descending from $u$.

Given the $i$-th instance in the dataset, we use the same symbol $x_i$ to denote the inputs to both the teacher and student models for notational convenience. The classifier produces a taxonomic hierarchical assignment, and we denote by $y_i \in \mathcal{V}$ the finest-grained label in the hierarchical structure defined by the classifier output. The model, parameterized by $\theta$, receives $x_i$ and outputs a logit vector for all leaf nodes, denoted as $\mathbf{z}_i \in \mathbb{R}^{|\mathcal{N}|}$, where $z_{i, l}$ represents the $l$-th element of the vector. To mitigate the risk of false positive classifications, the classification path extends exclusively to child nodes with a probability exceeding 0.5, and is ultimately truncated using a strict threshold of 0.80.

\subsection{TaxDistill Framework}
\label{sec:taxdistill_framework}

The base probability of a leaf node $l \in \mathcal{N}$ is computed via the standard softmax function:
\begin{equation}
P(l|x_i; \theta) = \frac{\exp(z_{i, l})}{\sum_{k \in \mathcal{N}} \exp(z_{i, k})}.
\end{equation}

Building upon this, the probability of any non-leaf node $u$ is defined as the marginalized sum of the probabilities of all its leaf descendants:
\begin{equation}
P(u|x_i; \theta) = \sum_{l \in leaves(u)} P(l|x_i; \theta).
\end{equation}

For the $i$-th sample, let $\mathcal{P}(y_i)$ be its initial hierarchical label path (i.e., the set of all nodes from the root to the assigned node $y_i$). The deep hierarchical loss maximizes the joint log-likelihood of all nodes along this path over a batch of size $N$:
\begin{equation}
\mathcal{L}_{\text{hier}}(\theta) = - \frac{1}{N} \sum_{i=1}^{N} \sum_{u \in \mathcal{P}(y_i)} \log P(u|x_i; \theta).
\end{equation}

The temperature parameter controls the smoothness of the output distribution; as the temperature increases, the distribution becomes more uniform, thereby emphasizing information from non-dominant classes. We introduce a temperature scaling parameter $\tau > 1$ to soften the probability distributions. Let $\theta_T$ and $\theta_S$ parameterize the teacher and student models, respectively, with their output logits denoted as $\mathbf{z}_i^T$ and $\mathbf{z}_i^S$. The softened distribution is defined as:
\begin{equation}
q^M(l|\mathbf{z}_i^M; \tau) = \frac{\exp(z_{i, l}^M / \tau)}{\sum_{k \in \mathcal{N}} \exp(z_{i, k}^M / \tau)}, \quad M \in \{S, T\}.
\end{equation}

During joint training, to prevent the randomly initialized student model from deviating from the teacher's probability distribution, we apply a stop-gradient ($\text{sg}$) operation to the teacher's logits. This ensures that the teacher is updated exclusively by the hierarchical loss derived from the initial labels. We then minimize the Kullback-Leibler (KL) divergence between the student's prediction and the teacher's stop-gradient soft distribution:
\begin{equation}
\begin{split}
\mathcal{L}_{\text{KD}}(\theta_S) &= \frac{\tau^2}{N} \sum_{i=1}^{N} \sum_{l \in \mathcal{N}} q^T(l|\text{sg}(\mathbf{z}_i^T); \tau) \\
&\quad \times \log \left( \frac{q^T(l|\text{sg}(\mathbf{z}_i^T); \tau)}{q^S(l|\mathbf{z}_i^S; \tau)} \right).
\end{split}
\end{equation}

Ultimately, the optimization objective for the teacher model $\theta_T$ is to minimize its own hierarchical supervision loss:
\begin{equation}
\mathcal{L}_{\text{Teacher}}(\theta_T) = \mathcal{L}_{\text{hier}}(\theta_T).
\end{equation}

Conversely, the optimization objective for the student model $\theta_S$ is a weighted combination of the hierarchical hard-label loss and the soft-label distillation loss, balanced by a hyperparameter $\alpha$:
\begin{equation}
\mathcal{L}_{\text{Student}}(\theta_S) = \alpha \mathcal{L}_{\text{hier}}(\theta_S) + (1 - \alpha) \mathcal{L}_{\text{KD}}(\theta_S).
\end{equation}

The total loss for the end-to-end framework is the sum of both components:
\begin{equation}
\mathcal{L}_{\text{total}} = \mathcal{L}_{\text{Teacher}}(\theta_T) + \mathcal{L}_{\text{Student}}(\theta_S).
\end{equation}

\section{Experiments}
\label{sec:experiments}

Following Taxometer, TaxDistill adopts a target-specific, on-the-fly recalibration protocol. For each newly analyzed microbial community, retrieval-based tools first generate initial taxonomic assignments, and the post-hoc model then corrects these predictions using only target-sample features, abundance profiles, the local taxonomic structure, and noisy pseudo-labels. Thus, our goal is not to train a universal cross-environment classifier, but to recalibrate existing retrieval outputs under substantial ecological and taxonomic shifts. CAMI2 ground-truth annotations are used only for evaluation and are never used during optimization.

\begin{table*}[t]
\centering
% ACM 规范：表格的 caption 和 label 必须放在表格正上方
\caption{Comprehensive evaluation of post-correction models across diverse environmental datasets. The table reports the F1 score, Recall (Rec.), and Precision (Prec.) achieved by three base classifiers before and after applying various post-correction strategies. The total number of contigs ($N$) for each dataset is indicated in the first column.}
\label{tab1}

\resizebox{\textwidth}{!}{
\begin{tabular}{ll rrr rrr rrr}
\toprule
\multirow{2}{*}{\textbf{Dataset}} & \multirow{2}{*}{\textbf{Correction Models}} & \multicolumn{3}{c}{\textbf{Metabuli}} & \multicolumn{3}{c}{\textbf{MMseqs2}} & \multicolumn{3}{c}{\textbf{Kraken2}} \\
\cmidrule(lr){3-5} \cmidrule(lr){6-8} \cmidrule(lr){9-11}
& & F1 score & Rec. & Prec. & F1 score & Rec. & Prec. & F1 score & Rec. & Prec. \\
\midrule

% ==================== 第 1 个数据集 ====================
\multirow{5}{*}{\begin{tabular}[c]{@{}l@{}}\textbf{\textit{Gastrointestinal}}\\ (N=81,602)\end{tabular}} 
& Base                     & 0.9086 & 0.8825 & 0.9363 & 0.7631 & 0.6380 & 0.9494 & \textbf{0.8533} & 0.8298 & 0.8781 \\
& Taxometer                & 0.9293 & 0.9177 & 0.9413 & 0.9237 & 0.9112 & 0.9366 & 0.8432 & 0.8031 & 0.8874 \\
& Taxometer\_G             & \textbf{0.9356} & 0.9188 & 0.9531 & 0.9298 & 0.9097 & 0.9508 & 0.8457 & 0.7989 & 0.8982 \\
& TaxDistill (100M)        & 0.9348 & 0.9246 & 0.9451 & 0.9408 & 0.9308 & 0.9511 & 0.8500 & 0.8161 & 0.8868 \\
& TaxDistill (500M)        & 0.9346 & 0.9250 & 0.9444 & \textbf{0.9412} & 0.9310 & 0.9517 & 0.8513 & 0.8168 & 0.8888 \\
\midrule

% ==================== 第 2 个数据集 ====================
\multirow{5}{*}{\begin{tabular}[c]{@{}l@{}}\textbf{\textit{Airways}}\\ (N=187,685)\end{tabular}} 
& Base                     & 0.9044 & 0.8781 & 0.9323 & 0.7542 & 0.6243 & 0.9523 & \textbf{0.8765} & 0.8446 & 0.9109 \\
& Taxometer                & 0.9145 & 0.8978 & 0.9318 & 0.9048 & 0.8787 & 0.9324 & 0.8618 & 0.8152 & 0.9141 \\
& Taxometer\_G             & 0.9081 & 0.8822 & 0.9357 & 0.8952 & 0.8601 & 0.9334 & 0.8540 & 0.8003 & 0.9154 \\
& TaxDistill (100M)        & 0.9214 & 0.8921 & 0.9526 & 0.9198 & 0.8878 & 0.9542 & 0.8497 & 0.7909 & 0.9180 \\
& TaxDistill (500M)        & \textbf{0.9215} & 0.8925 & 0.9524 & \textbf{0.9205} & 0.8887 & 0.9546 & 0.8517 & 0.7937 & 0.9187 \\
\midrule

% ==================== 第 3 个数据集 ====================
\multirow{5}{*}{\begin{tabular}[c]{@{}l@{}}\textbf{\textit{Skin}}\\ (N=173,927)\end{tabular}} 
& Base                     & 0.9055 & 0.8797 & 0.9328 & 0.7517 & 0.6207 & 0.9527 & \textbf{0.8891} & 0.8676 & 0.9118 \\
& Taxometer                & 0.9080 & 0.8813 & 0.9364 & 0.9055 & 0.8788 & 0.9338 & 0.8770 & 0.8426 & 0.9143 \\
& Taxometer\_G             & 0.9067 & 0.8745 & 0.9414 & 0.8936 & 0.8575 & 0.9329 & 0.8668 & 0.8235 & 0.9149 \\
& TaxDistill (100M)        & 0.9199 & 0.8838 & 0.9591 & \textbf{0.9177} & 0.8830 & 0.9552 & 0.8678 & 0.8219 & 0.9191 \\
& TaxDistill (500M)        & \textbf{0.9201} & 0.8834 & 0.9599 & 0.9173 & 0.8836 & 0.9537 & 0.8669 & 0.8208 & 0.9184 \\
\midrule

% ==================== 第 4 个数据集 ====================
\multirow{5}{*}{\begin{tabular}[c]{@{}l@{}}\textbf{\textit{Oral}}\\ (N=201,563)\end{tabular}} 
& Base                     & 0.8991 & 0.8722 & 0.9277 & 0.7414 & 0.6098 & 0.9454 & \textbf{0.8919} & 0.8774 & 0.9069 \\
& Taxometer                & 0.9130 & 0.8955 & 0.9313 & 0.9085 & 0.8876 & 0.9304 & 0.8880 & 0.8647 & 0.9126 \\
& Taxometer\_G             & 0.9078 & 0.8823 & 0.9348 & 0.9005 & 0.8712 & 0.9318 & 0.8827 & 0.8535 & 0.9140 \\
& TaxDistill (100M)        & 0.9227 & 0.8973 & 0.9496 & 0.9224 & 0.8966 & 0.9498 & 0.8890 & 0.8622 & 0.9175 \\
& TaxDistill (500M)        & \textbf{0.9237} & 0.8988 & 0.9500 & \textbf{0.9228} & 0.8978 & 0.9492 & 0.8890 & 0.8616 & 0.9181 \\
\midrule

% ==================== 第 5 个数据集 ====================
\multirow{5}{*}{\begin{tabular}[c]{@{}l@{}}\textbf{\textit{Urogenital}}\\ (N=57,529)\end{tabular}} 
& Base                     & 0.8987 & 0.8703 & 0.9290 & 0.7486 & 0.6193 & 0.9461 & 0.8846 & 0.8787 & 0.8906 \\
& Taxometer                & 0.9026 & 0.8851 & 0.9207 & 0.8893 & 0.8670 & 0.9129 & 0.8800 & 0.8672 & 0.8932 \\
& Taxometer\_G             & 0.8980 & 0.8760 & 0.9211 & 0.8789 & 0.8486 & 0.9115 & 0.8805 & 0.8669 & 0.8946 \\
& TaxDistill (100M)        & 0.9097 & 0.8915 & 0.9287 & 0.8940 & 0.8763 & 0.9123 & \textbf{0.8857} & 0.8763 & 0.8953 \\
& TaxDistill (500M)        & \textbf{0.9103} & 0.8927 & 0.9287 & \textbf{0.8942} & 0.8761 & 0.9130 & 0.8855 & 0.8760 & 0.8952 \\
\midrule

% ==================== 第 6 个数据集 ====================
\multirow{5}{*}{\begin{tabular}[c]{@{}l@{}}\textbf{\textit{Rhizosphere}}\\ (N=28,318)\end{tabular}} 
& Base                     & 0.7854 & 0.7809 & 0.7899 & 0.6936 & 0.5863 & 0.8488 & 0.6122 & 0.5694 & 0.6619 \\
& Taxometer                & 0.7857 & 0.7813 & 0.7901 & 0.7720 & 0.7664 & 0.7777 & 0.6466 & 0.5659 & 0.7541 \\
& Taxometer\_G             & 0.7851 & 0.7796 & 0.7908 & 0.7738 & 0.7659 & 0.7819 & 0.6496 & 0.5653 & 0.7635 \\
& TaxDistill (100M)        & 0.7919 & 0.7757 & 0.8089 & \textbf{0.7889} & 0.7708 & 0.8078 & \textbf{0.6743} & 0.5646 & 0.8367 \\
& TaxDistill (500M)        & \textbf{0.7924} & 0.7755 & 0.8099 & 0.7882 & 0.7706 & 0.8065 & 0.6728 & 0.5645 & 0.8325 \\
\midrule

% ==================== 第 7 个数据集 ====================
\multirow{5}{*}{\begin{tabular}[c]{@{}l@{}}\textbf{\textit{Marine}}\\ (N=216,508)\end{tabular}} 
& Base                     & 0.8394 & 0.8246 & 0.8548 & 0.7439 & 0.6526 & 0.8650 & 0.7455 & 0.7104 & 0.7842 \\
& Taxometer                & 0.8483 & 0.8353 & 0.8617 & 0.8472 & 0.8329 & 0.8621 & 0.7546 & 0.7008 & 0.8174 \\
& Taxometer\_G             & 0.8479 & 0.8300 & 0.8667 & 0.8430 & 0.8232 & 0.8638 & 0.7581 & 0.7012 & 0.8250 \\
& TaxDistill (100M)        & 0.8654 & 0.8350 & 0.8981 & 0.8635 & 0.8434 & 0.8846 & 0.7657 & 0.6998 & 0.8451 \\
& TaxDistill (500M)        & \textbf{0.8660} & 0.8354 & 0.8990 & \textbf{0.8648} & 0.8453 & 0.8853 & \textbf{0.7673} & 0.7018 & 0.8463 \\
\bottomrule
\end{tabular}
}
\end{table*}

\begin{figure*}[t]
    \centering
    \includegraphics[width=\textwidth]{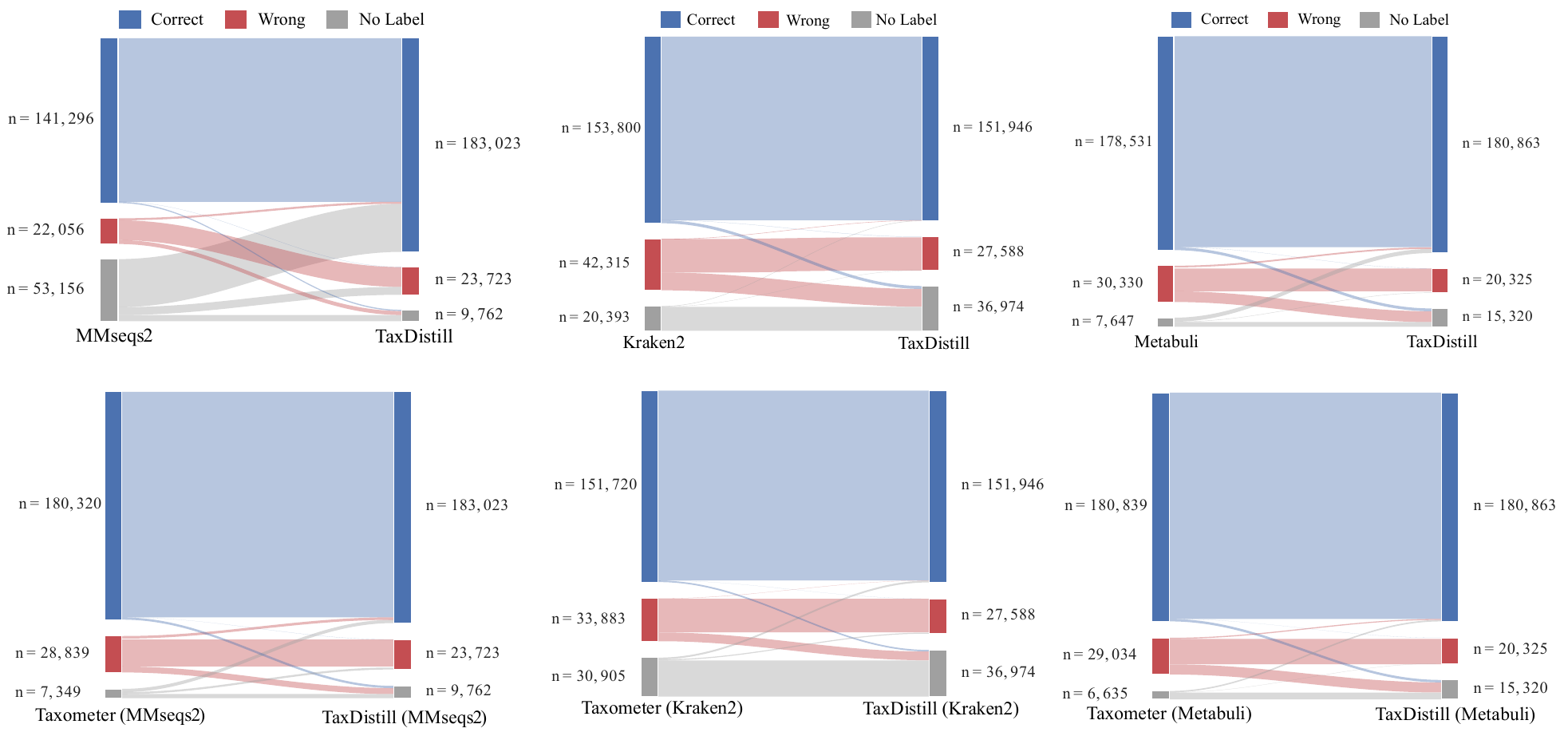}
    \caption{Sankey diagram analysis of label transition and recalibration dynamics on the Marine dataset. The first row compares initial labeling tools with their TaxDistill optimized versions, while the second row contrasts optimized Taxometer with TaxDistill.}
    \label{Figure3}
\end{figure*}

\subsection{Baseline Settings}
\label{subsec:baselines_setting}

We systematically evaluate the proposed TaxDistill framework on the widely adopted CAMI2 benchmark datasets \citep{meyer2022critical} and comprehensively compare its performance against mainstream metagenomic classification baseline models. The detailed dataset processing procedures and model evaluation metrics are provided in the Appendix \ref{sec:exp_details}.

TaxDistill is designed as a flexible post-hoc label correction framework. To validate its performance, we evaluate it against a hierarchical set of baselines, divided into initial retrieval tools and post-hoc correction models.

For initial retrieval baselines, we use three mainstream retrieval tools to generate base predictions: 1) \textbf{MMseqs2 (v18.8cc5c)} \citep{kallenborn2025gpu} for $k$-mer pre-screening homology search; 2) \textbf{Metabuli (v1.1.1)} \citep{kim2024metabuli} for bimodal LCA alignment; and 3) \textbf{Kraken2 (v2.17.1)} \citep{wood2019improved} for exact $k$-mer matching.

Post-hoc correction baselines. We subsequently apply denoising frameworks to refine the initial predictions. Specifically, we distill knowledge from pre-trained GenomeOcean models with different scales (100M and 500M parameters) as teacher models, resulting in two variants of TaxDistill, namely \textbf{TaxDistill (100M)} and \textbf{TaxDistill (500M)}. We compare TaxDistill against three experimental setups: 1) the raw uncorrected outputs of the initial tools; 2) \textbf{Taxometer} \citep{kutuzova2024taxometer}, which recalibrates using TNFs and sample abundances; and 3) \textbf{Taxometer\_G}, our modified version of Taxometer, in which the original TNFs are replaced with embeddings derived solely from GenomeOcean to evaluate the impact of foundation model features.

\subsection{Results on Human Microbiome Datasets}
\label{subsec:cami2_human}

\begin{table*}[t]
\centering
% ACM 规范：表格的 caption 和 label 必须放在表格正上方
\caption{Performance comparison of GenomeOcean and TaxDistill across Marine and Urogenital datasets.}
\label{tab2}

\resizebox{\textwidth}{!}{ 
\begin{tabular}{ll rrr rrr rrr}
\toprule
\multirow{2}{*}{\textbf{Dataset}} & \multirow{2}{*}{\textbf{Correction Models}} & \multicolumn{3}{c}{\textbf{Metabuli}} & \multicolumn{3}{c}{\textbf{MMseqs2}} & \multicolumn{3}{c}{\textbf{Kraken2}} \\
\cmidrule(lr){3-5} \cmidrule(lr){6-8} \cmidrule(lr){9-11}
& & F1 score & Rec. & Prec. & F1 score & Rec. & Prec. & F1 score & Rec. & Prec. \\
\midrule

% ==================== Marine 数据集 ====================
\multirow{4}{*}{\begin{tabular}[c]{@{}l@{}}\textbf{\textit{Marine}}\\ (N=216,508)\end{tabular}} 
& GenomeOcean (100M)       & 0.7170 & 0.6279 & 0.8355 & 0.7169 & 0.6352 & 0.8227 & 0.6125 & 0.5017 & 0.7861 \\
& GenomeOcean (500M)       & 0.7203 & 0.6347 & 0.8327 & 0.7200 & 0.6416 & 0.8203 & 0.7203 & 0.6347 & 0.8327 \\
& TaxDistill (100M)        & 0.8654 & 0.8350 & 0.8981 & 0.8635 & 0.8434 & 0.8846 & 0.7657 & 0.6998 & 0.8451 \\
& TaxDistill (500M)        & \textbf{0.8660} & 0.8354 & 0.8990 & \textbf{0.8648} & 0.8453 & 0.8853 & \textbf{0.7673} & 0.7018 & 0.8463 \\
\midrule

% ==================== Urogenital 数据集 ====================
\multirow{4}{*}{\begin{tabular}[c]{@{}l@{}}\textbf{\textit{Urogenital}}\\ (N=57,529)\end{tabular}} 
& GenomeOcean (100M)       & 0.7320 & 0.6235 & 0.8861 & 0.7327 & 0.6245 & 0.8861 & 0.6975 & 0.6044 & 0.8245 \\
& GenomeOcean (500M)       & 0.7326 & 0.6207 & 0.8937 & 0.7324 & 0.6206 & 0.8934 & 0.6970 & 0.6010 & 0.8295 \\
& TaxDistill (100M)        & 0.9097 & 0.8915 & 0.9287 & 0.8940 & 0.8763 & 0.9123 & \textbf{0.8857} & 0.8763 & 0.8953 \\
& TaxDistill (500M)        & \textbf{0.9103} & 0.8927 & 0.9287 & \textbf{0.8942} & 0.8761 & 0.9130 & 0.8855 & 0.8760 & 0.8952 \\
\bottomrule
\end{tabular}
}
\end{table*}

We systematically evaluated the proposed TaxDistill framework and baseline models on five highly diverse CAMI2 human microbiome datasets. As shown in Table \ref{tab1} and Table \ref{tabs1}, we report the F1 score alongside the number of correctly classified (\textit{Correct}), misclassified (\textit{Wrong}), and unassigned (\textit{No label}) sequences. Across 15 scenarios from five datasets and three classification tools, TaxDistill outperformed Taxometer in 13 cases and showed effective confidence-based recalibration, reassigning many Taxometer errors to either correct labels or the unclassified category.

On the Gastrointestinal dataset using MMseqs2, Taxometer improves the initial F1 score from 0.763 to 0.924, while TaxDistill further increases it to 0.941. Similarly, on the Oral and Skin datasets using Metabuli, TaxDistill consistently outperforms Taxometer by about 1.0\% to 1.2\% in F1 score. This indicates that, compared to the hand-crafted shallow features used by Taxometer, the deep semantic knowledge from genomic foundation models enables the network to perform more accurate metagenomic taxonomic annotation.

Experimental results indicate that MMseqs2 and Metabuli exhibit relatively conservative predictions. When employed to generate initial pseudo-labels, these base classifiers yield a substantial number of unassigned contigs. While Taxometer can partially recover these sequences, TaxDistill shows a stronger rescue capability. For instance, on the Airways dataset using MMseqs2, Taxometer achieves 164,925 correct taxonomic assignments. TaxDistill further improves upon this result, increasing the correct classifications to 166,804. Overall, TaxDistill reduces the total number of misclassified sequences by nearly 4,000, of which approximately 2,000 sequences are corrected from previously incorrect assignments. 

In contrast, Kraken2 uses an aggressive exact $k$-mer matching strategy that tends to force classifications. This results in fewer unassigned sequences but introduces a large number of false-positive errors. Under this heavy label noise, Taxometer shows a relatively conservative calibration, predicting more unclassified classes, whereas TaxDistill applies an even stricter calibration strategy. Taking the Airways dataset as an example, the base Kraken2 outputs 13,670 unassigned labels and 15,499 erroneous predictions. Taxometer increases the unassigned labels to 20,300 and reduces the errors to 14,377. TaxDistill applies stronger distillation regularization, significantly increasing the unclassified labels to 25,528 while further reducing the errors to 13,183. Although this conservative strategy of converting high-confidence errors into unassigned labels may slightly lower the F1 score on a few datasets compared to Taxometer, avoiding false positives is often much more critical than forcing a prediction in clinical metagenomic applications.

\subsection{Rhizosphere and Marine Datasets}
\label{subsec:rhizosphere_marine}

To further validate the label correction capability of TaxDistill in complex real-world scenarios, we conducted an in-depth analysis on two highly challenging environmental datasets: Rhizosphere and Marine. Unlike microbiomes associated with humans, the marine dataset contains a vast amount of uncultured microbial genomic fragments and plasmids. Meanwhile, the Rhizosphere dataset represents one of the most diverse and structurally complex ecosystems on Earth.

As shown in Table \ref{tab1} and Table \ref{tabs1}, when processing the marine dataset rich in uncultured fragments, TaxDistill demonstrates superior classification performance over Taxometer. The performance improvements are particularly notable when using MMseqs2 and Metabuli as baselines. Taking MMseqs2 as an example, Taxometer improves the base F1 score from 0.743 to 0.847, whereas TaxDistill further elevates it to 0.864. Similarly, TaxDistill yields a steady improvement of approximately 1.5\% on both Metabuli and Kraken2. This demonstrates that Taxometer, which relies on hand-crafted features, is prone to encountering representational bottlenecks. 

On the highly complex Rhizosphere dataset, under the MMseqs2 baseline in Table \ref{tab1}, Taxometer attempts to recover a large number of unassigned sequences. Although this pushes its F1 score to 0.772, the number of misclassifications also expands significantly. In contrast, TaxDistill adopts a more conservative confidence-based recalibration strategy. When faced with highly ambiguous subspecies boundaries, it is more capable than Taxometer of assigning potentially unreliable predictions to the unclassified category, thereby reducing the absolute number of misclassifications. This trend is equally evident with the Metabuli baseline. Notably, for the Kraken2 classifier, TaxDistill successfully reassigns a large number of erroneous predictions to the unclassified category. Compared to Taxometer, TaxDistill achieves an approximate 2.6\% improvement in the F1 score and a striking 8\% surge in precision, while maintaining a roughly equivalent recall. This demonstrates that in complex environments, TaxDistill can control false positives for aggressive classifiers through conservative confidence-based recalibration.

Overall, across 21 test scenarios covering five human microbiome datasets, the Rhizosphere dataset, the Marine dataset, and three classification tools, TaxDistill outperformed Taxometer\_G in 19 cases. This consistent trend suggests that simply using GenomeOcean embeddings as replacement features is insufficient to fully exploit foundation-model knowledge. In contrast, explicit distillation provides instance-specific teacher guidance that more effectively regularizes the student under noisy pseudo-label supervision. In addition, the 500M teacher model generally achieves slightly stronger label correction performance than the 100M model.

\subsection{Sankey Diagram Analysis}

To further analyze the recalibration mechanism, we visualize the label transition process on the Marine dataset using Sankey diagrams in Figure \ref{Figure3}. First, compared to the initial heuristic classifiers, TaxDistill achieves a substantial performance gain. Using MMseqs2 as an example, the number of correct classifications surges from 141,296 to 183,023. The Sankey flow illustrates that TaxDistill not only precisely redirects a large number of unassigned contigs to the correct category, but also effectively mitigates initial misclassifications: a portion is directly corrected, while the remainder is conservatively assigned to the unclassified category. For baselines like Kraken2 and Metabuli, TaxDistill maintains recalibration gains while showing a propensity to convert initial errors into unclassified labels. 

Furthermore, in comparison with the advanced baseline Taxometer, TaxDistill demonstrates a highly favorable recalibration trade-off. Although it conservatively reassigns a negligible fraction of Taxometer's correct predictions to the unclassified state, it successfully converts a larger number of erroneous and unassigned predictions into correct labels, and is also able to assign erroneous predictions to the unclassified category. Notably, on the Metabuli baseline, TaxDistill not only yields more correct predictions than Taxometer but also significantly reduces the number of misclassifications by nearly 9,000. This further confirms the effective recalibration capability of our model.

\section{Why Not Use the Teacher Model Directly?}

A natural question is why TaxDistill uses GenomeOcean as a teacher rather than directly adopting it as the final taxonomic classifier. As shown in Table~\ref{tab2}, GenomeOcean with a fine-tuned classification head performs substantially worse than TaxDistill on both the Marine and Urogenital datasets. This result is consistent with the design motivation of TaxDistill: the genomic foundation model is used not as an oracle classifier, but as a semantic regularizer for lightweight post-hoc recalibration.

The performance gap mainly arises from two practical constraints. First, GenomeOcean has a maximum input length and therefore cannot always process full contig sequences. Long contigs must be truncated before being passed to the teacher, which discards sequence context beyond the truncation boundary and weakens taxonomic discrimination based on long-range genomic signals. In contrast, the student branch uses full-contig-derived features represented as a $(103 + K + 1)$-dimensional vector, allowing it to exploit information from the complete contig and mitigate the systematic information loss caused by teacher-side truncation.

Second, we only fine-tune the classification head of GenomeOcean, keeping its pre-trained backbone frozen. This preserves the general genomic knowledge learned during large-scale pre-training, but it also limits the ability of GenomeOcean to adapt its representation space to dataset-specific taxonomic structures. More aggressive adaptation strategies, such as parameter-efficient fine-tuning, may improve the direct classifier but would increase training and deployment costs, which is undesirable for practical metagenomic analysis scenarios such as rapid pathogen detection.

These limitations do not prevent GenomeOcean from serving as an effective teacher. Knowledge distillation does not require the teacher to be strictly superior in top-1 accuracy; even a structurally limited teacher can provide useful supervision through its output distribution \cite{lee2023study, lu2026strong}. The key value of GenomeOcean lies in the relational information encoded in its soft predictions. Unlike one-hot hard labels, which impose a single target and identical penalties on all non-target classes, the teacher's softened outputs provide a continuous distribution over candidate taxa. Even when truncation leads to an incorrect top-1 prediction, this distribution can still preserve taxonomic similarities and hierarchical semantic relationships learned during large-scale genomic pre-training.

Recent analyses connect knowledge distillation with label smoothing, suggesting that distillation acts as a learned, instance-specific regularizer \cite{yuan2020revisiting, muller2019does}. In TaxDistill, this regularization helps the student avoid overfitting noisy hard pseudo-labels, suppresses overconfident predictions, and encourages smoother decision boundaries. Therefore, the goal of TaxDistill is not to replicate GenomeOcean's final predictions. Instead, it combines the relational genomic knowledge provided by the teacher with the full-contig information available to the student, enabling a lightweight model to achieve more reliable taxonomic recalibration than either the noisy retrieval output or GenomeOcean with a fine-tuned classification head.

\section{Conclusion}

In this paper, we presented TaxDistill, a plug-and-play knowledge distillation framework for reliable metagenomic taxonomic recalibration under noisy pseudo-labels. TaxDistill is motivated by a practical challenge in complex microbiome analysis: retrieval-based classifiers are efficient, but reference incompleteness and microbial novelty frequently lead to erroneous assignments and unassigned contigs. Rather than replacing these classifiers, TaxDistill acts as a post-hoc correction tool that recalibrates their outputs using GenomeOcean as a semantic teacher.

Our analysis further shows why directly using GenomeOcean with a fine-tuned classification head is not the optimal solution. Although GenomeOcean captures rich genomic semantics through large-scale pre-training, its direct use as a classifier is limited by input-length truncation and by the restricted adaptation of a frozen backbone with a fine-tuned classification head. TaxDistill resolves this tension by using GenomeOcean not as an oracle classifier, but as a relational regularizer. The teacher's softened taxonomic outputs transfer inter-taxon relationships and hierarchical semantic structure, while the lightweight student leverages full-contig-derived features to compensate for teacher-side truncation and reduce overfitting to noisy hard labels.

Experiments on seven diverse CAMI2 datasets demonstrate that this design improves annotation reliability over retrieval-based classifiers and the Taxometer calibration baseline in most scenarios. In both human-associated and environmental microbiomes, TaxDistill recovers unassigned contigs while reducing false-positive assignments through confidence-based conservative recalibration. These results close the loop between foundation-model knowledge and practical metagenomic annotation: large genomic models need not serve as final classifiers to be useful; when distilled appropriately, they can provide reliable semantic guidance for lightweight recalibration in ecological monitoring and clinical metagenomic analysis.

\section{Limitations and Ethical Considerations}
Although TaxDistill improves metagenomic label correction, two limitations remain. First, TaxDistill is constrained by the teacher model and may inherit biases from GenomeOcean, reference databases, and upstream retrieval-based classifiers, leading to less reliable annotations for underrepresented or poorly curated taxa. Second, TaxDistill relies on conservative rejection by assigning potentially unreliable predictions to the unclassified category; thus, it should be viewed as a post-hoc correction tool for reference-supported taxa rather than a method for de novo discovery of unknown species.

From an ethical perspective, the main concern is the downstream use of taxonomic annotations in sensitive biomedical, ecological, or public-health contexts. Incorrect annotations may lead to misleading biological interpretation if used as standalone evidence. TaxDistill is therefore intended as a decision-support component rather than an autonomous diagnostic or surveillance system, and its predictions should be interpreted together with domain expertise, database provenance, and additional experimental validation.

\section{Generative AI Usage}
Generative AI tools were used only for language polishing and grammar checking. They were not used to generate experimental results, analyze data, or make scientific claims. All technical content, experiments, and conclusions were verified by the authors.

\section{Author Contributions}
R.Y. and L.L. led the artificial intelligence methodology, including the design of the TaxDistill framework, experimental analysis, and manuscript writing. S.S. supervised the study, provided biological guidance for problem formulation and critically revised the manuscript. Y.Z. and Z.L. contributed biological insights, assisted with metagenomic data interpretation. Z.Z. provided writing guidance and helped improve the manuscript presentation.

\bibliographystyle{ACM-Reference-Format}
\bibliography{custom}

\appendix
\section{Appendix}
\label{sec:appendix}

\subsection{Code and Dataset Availability}

Our data can be accessed at 
\href{https://cami-challenge.org/datasets/}{CAMI Challenge datasets}, 
and the code is released at \url{https://anonymous.4open.science/r/TaxDistill-zzzz}.

\subsection{Experimental Details}
\label{sec:exp_details}

\textbf{Metric Computation.} The evaluation metrics used in this study are defined as follows:

\begin{equation}
\text{Recall} = \frac{\mathit{Correct}}{\mathit{Correct} + \mathit{Wrong} + \mathit{No\ label}},
\end{equation}

\begin{equation}
\text{Precision} = \frac{\mathit{Correct}}{\mathit{Correct} + \mathit{Wrong}}.
\end{equation}

\noindent\textbf{Dataset Settings.} To ensure a fair comparison, we use the same benchmark datasets and preprocessing pipeline as Taxometer. Specifically, to systematically validate the performance of the TaxDistill framework against noisy pseudo-labels and its self-correction capabilities, we construct a comprehensive benchmark suite primarily sourced from the CAMI2 datasets. This benchmark encompasses highly diverse microbial environments, including five human microbiome datasets and two complex environmental microbiome datasets. All performance evaluations are conducted at the species level.

\begin{table*}[t]
\centering
% ACM 规范：表格的 caption 和 label 必须放在表格正上方
\caption{Ablation study on the Marine Dataset. The table reports the performance variations (F1 score, Recall, and Precision) across three different tools under different $\alpha$ ratios (top block) and temperature settings (bottom block).}
\label{tab3}

\resizebox{\textwidth}{!}{ 
\begin{tabular}{l ccc ccc ccc}
\toprule
\multirow{2}{*}{\textbf{Settings}} & \multicolumn{3}{c}{\textbf{MMseqs2}} & \multicolumn{3}{c}{\textbf{Metabuli}} & \multicolumn{3}{c}{\textbf{Kraken2}} \\
\cmidrule(lr){2-4} \cmidrule(lr){5-7} \cmidrule(lr){8-10}
& F1 score & Recall & Prec. & F1 score & Recall & Prec. & F1 score & Recall & Prec. \\
\midrule

$\alpha=0.3$ & 0.8648 & 0.8453 & 0.8853 & 0.8660 & 0.8354 & 0.8990 & 0.7673 & 0.7018 & 0.8463 \\
$\alpha=0.4$ & 0.8644 & 0.8459 & 0.8838 & 0.8649 & 0.8358 & 0.8962 & 0.7664 & 0.7028 & 0.8428 \\
$\alpha=0.5$ & 0.8635 & 0.8453 & 0.8825 & 0.8640 & 0.8370 & 0.8928 & 0.7662 & 0.7036 & 0.8410 \\
$\alpha=0.6$ & 0.8625 & 0.8441 & 0.8817 & 0.8626 & 0.8369 & 0.8899 & 0.7647 & 0.7041 & 0.8366 \\
$\alpha=0.7$ & 0.8620 & 0.8443 & 0.8805 & 0.8606 & 0.8363 & 0.8863 & 0.7646 & 0.7042 & 0.8362 \\
$\alpha=0.8$ & 0.8607 & 0.8416 & 0.8806 & 0.8588 & 0.8371 & 0.8817 & 0.7640 & 0.7048 & 0.8342 \\
\midrule
$T=3$ & 0.8635 & 0.8376 & 0.8911 & 0.8634 & 0.8303 & 0.8993 & 0.7669 & 0.6955 & 0.8546 \\
$T=4$ & 0.8648 & 0.8453 & 0.8853 & 0.8660 & 0.8354 & 0.8990 & 0.7673 & 0.7018 & 0.8463 \\
$T=5$ & 0.8653 & 0.8465 & 0.8849 & 0.8665 & 0.8369 & 0.8982 & 0.7666 & 0.7028 & 0.8430 \\
$T=6$ & 0.8653 & 0.8468 & 0.8846 & 0.8661 & 0.8366 & 0.8977 & 0.7668 & 0.7031 & 0.8433 \\
\bottomrule
\end{tabular}
}
\end{table*}

\begin{figure*}[t]
    \centering
    \includegraphics[width=\textwidth]{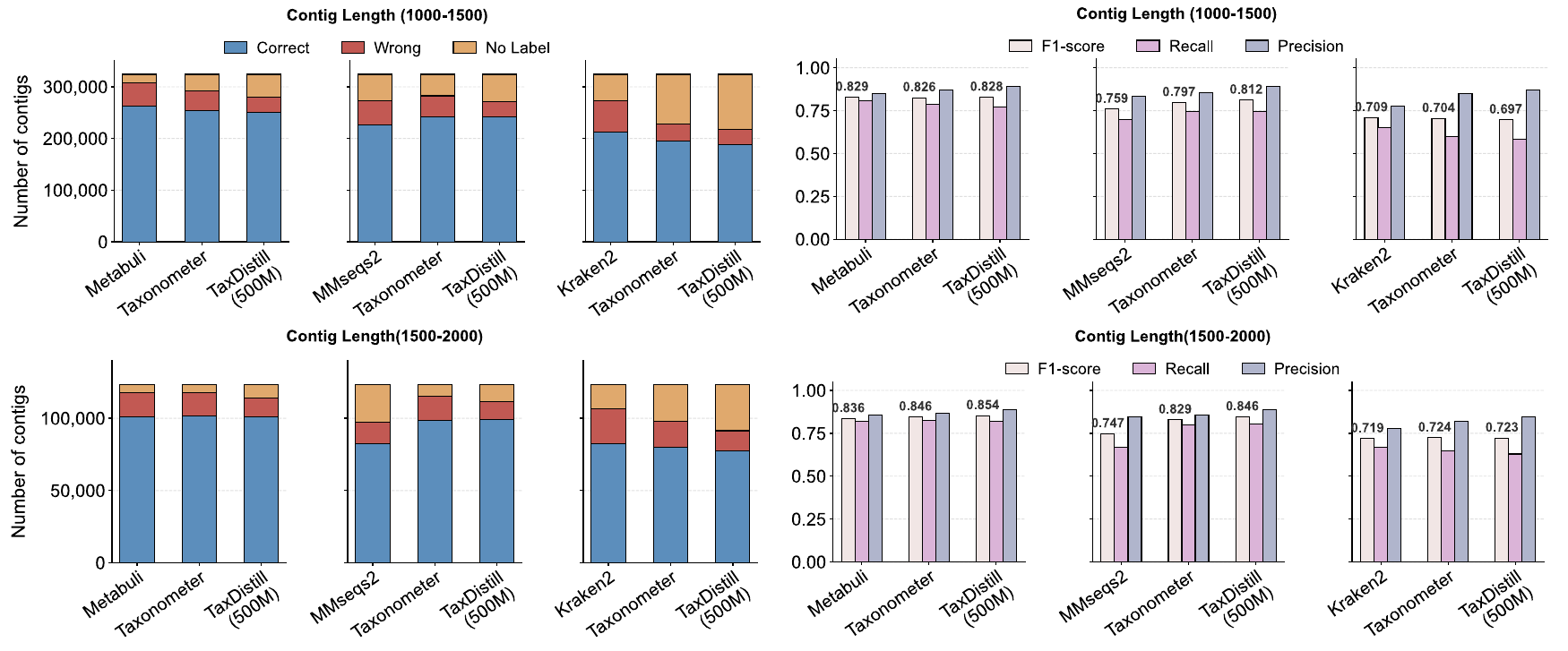}
    \caption{Ablation study on the effect of contig sequence length on model inference performance.}
    \label{Figure4}
\end{figure*}

\noindent\textbf{Sequence Assembly and Preprocessing.} Consistent with the analysis strategy of Taxometer, we conduct our evaluation using assembled sequences (contigs) rather than raw short reads. This strategy ensures that deep representation learning can capture sufficient contextual semantic information. We acquired the official contig and read files provided by the CAMI2 platform, which preserve the complete environmental characteristics of the metagenomic sequences. In the main experiments, we applied a strict length filtering criterion across all datasets, retaining only contigs with a length greater than or equal to $2,000$ base pairs (bp). Additionally, we evaluated the impact of varying assembly lengths on model performance in our ablation studies.

\noindent\textbf{Ground Truth Establishment and Label Alignment.} We followed Taxometer's label processing logic: for the five human microbiome datasets, we directly extracted species-level ground truth annotations using the GTDB-Tk tool, strictly adhering to the Genome Taxonomy Database (GTDB) nomenclature. However, the Marine and Rhizosphere datasets originally employed an NCBI-based taxonomic system. To standardize the label space across all datasets, we uniformly mapped the labels of these environmental datasets from the NCBI system to the GTDB (v226) system, and systematically discarded ambiguous sequences that lacked a strict one-to-one species-level mapping.

\noindent\textbf{Hyperparameters.} For a fair comparison, we adopt the same training configuration as Taxometer, setting the number of training epochs to 100 and the weight decay to $1\times10^{-4}$. For TaxDistill, the batch size is set to 64. The learning rate of the student model is kept consistent with Taxometer and set to $1\times10^{-3}$, while the teacher model uses a learning rate of $1\times10^{-4}$. For knowledge distillation, the distillation weight is set to 0.3 and the temperature is set to 4. Finally, we conduct ablation studies to systematically analyze the impact of these key hyperparameters.

\noindent\textbf{Hardware Details.} All experiments were conducted on a server equipped with two Intel(R) Xeon(R) Platinum 8558 CPUs (96 cores in total) and four NVIDIA GeForce RTX 5090 GPUs, each with 32 GB of memory. Although the server was equipped with four GPUs, the TaxDistill model requires only a single NVIDIA GeForce RTX 5090 GPU for both training; the remaining GPUs reflect the total hardware resources available on the server. All implementations were based on the PyTorch framework.

\subsection{Parameter Ablation Analysis}

To investigate the specific effects of knowledge distillation hyper-
parameters on model performance, we conducted a systematic
ablation study on the highly heterogeneous Marine dataset.

Regularization Effect of Distillation Weight ($\alpha$).
First, the distillation weight $\alpha$ determines the balance between the student model's reliance on the initial hard pseudo-labels and the teacher model's soft labels during joint optimization. Keeping other parameters constant, we evaluated the performance variations across $\alpha \in [0.3, 0.8]$. As shown in Table~\ref{tab3}, setting $\alpha = 0.3$ yields the optimal F1 score and precision across all three baseline classifiers. Conceptually, a lower $\alpha$ compels the student model to heavily leverage the probability distribution output by the teacher model as a strong regularizer. This mechanism effectively mitigates the student model's tendency to overfit the erroneous hard labels introduced by upstream heuristic tools, thereby enhancing overall performance.

Furthermore, with $\alpha$ fixed at 0.3, we analyzed the impact of the distillation temperature $T$. Increasing $T$ further softens the teacher model's logits, effectively smoothing the decision boundaries between ambiguous taxonomic classes. Experimental results indicate that while higher temperatures (e.g., $T \in \{5, 6\}$) marginally improve the overall F1 score by boosting recall, they inevitably lead to a degradation in precision. 
Given that controlling the risk of false positives is typically more critical than merely maximizing recall in complex real-world environmental samples, a careful balance is required. As a practical trade-off, setting $T=4$ sustains a highly competitive F1 score while effectively preventing the deterioration of precision.

\subsection{Contig Length Analysis}
In metagenomic analysis, contig sequences serve as the primary carriers of microbial genetic information; theoretically, longer sequences encapsulate richer contextual semantics. To investigate whether the model maintains robust label correction capabilities under information-constrained conditions, we conducted experimental analyses across two length intervals: 1000-1500 bp and 1500-2000 bp. As illustrated in Figure~\ref{Figure4}, on the MMseqs2 and Metabuli baselines, TaxDistill demonstrates consistently superior recalibration performance compared to Taxometer across both length intervals. This indicates that, benefiting from the effective transfer of global deep semantics facilitated by the knowledge distillation mechanism, TaxDistill can overcome the local feature deficits caused by low information density, thereby maintaining its recalibration advantage in short-sequence scenarios. On the Kraken2 classifier, although TaxDistill and Taxometer exhibit comparable overall F1 scores, TaxDistill achieves superior precision. When short sequences result in highly ambiguous classification boundaries, TaxDistill leverages confidence information from the output distribution to assign potentially unreliable predictions to the unclassified category, thereby improving precision.

\subsection{Time Analysis}

\begin{figure}[t] % 1. 这里去掉了星号
    \centering
    % 2. 这里将 \textwidth 改成了 \linewidth
    \includegraphics[width=\linewidth]{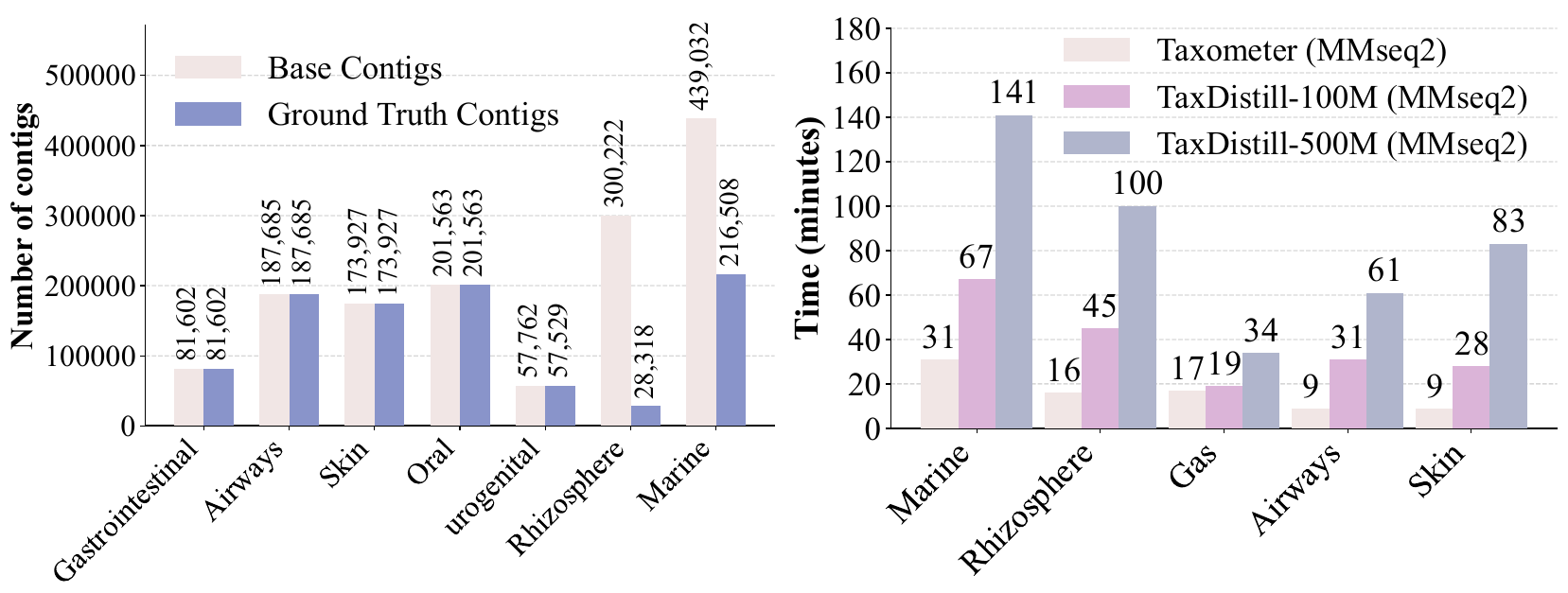}
    \caption{Dataset contig volume statistics and time overhead analysis for end-to-end inference. All experiments were conducted on a single NVIDIA GeForce RTX 5090 GPU.}
    \label{Figure5}
\end{figure} % 1. 结尾也要去掉星号

Finally, we systematically evaluated the computational efficiency and time overhead of the TaxDistill framework. As illustrated in the left panel of Figure~\ref{Figure5}, we report both the total number of contigs satisfying the initial length filtering criterion and the subset of contigs with ground-truth labels used for final performance evaluation across various datasets. The right panel of Figure~\ref{Figure5} provides a detailed comparison of the execution times between the Taxometer baseline and TaxDistill models of two different parameter scales across five datasets. 

Analysis of Figure~\ref{Figure5} reveals that while the introduction of a genomic foundation model for deep feature extraction increases inference time, the overall computational cost remains within a tractable range for practical applications. This efficiency is primarily attributed to the design of TaxDistill, which only requires fine-tuning a lightweight classification head. On the highly complex Marine dataset, the end-to-end runtime of TaxDistill (500M) is only approximately 110 minutes longer than that of the Taxometer baseline. Notably, the more lightweight TaxDistill (100M) variant introduces an additional overhead of only approximately 36 minutes on the Marine dataset. These results demonstrate that our framework improves recalibration performance while maintaining a manageable and practical time overhead.

% ==========================================
% 将下面这一行加在表格最前面，强制将其编号设为 S1
\setcounter{table}{0}
\renewcommand{\thetable}{S\arabic{table}}
% ==========================================

\begin{table*}[t]
\centering
% ACM 规范：Caption 和 Label 必须放在表格正上方
\caption{Comprehensive evaluation of post-correction models across diverse environmental datasets. The table reports the absolute number of Correct, Wrong, and Unlabeled (Unlbl.) contigs assigned by three base classifiers before and after applying various post-correction strategies. The total number of contigs ($N$) for each dataset is indicated in the first column.}
\label{tabs1}

\resizebox{\textwidth}{!}{ 
\begin{tabular}{ll rrr rrr rrr}
\toprule
\multirow{2}{*}{\textbf{Dataset}} & \multirow{2}{*}{\textbf{Correction Models}} & \multicolumn{3}{c}{\textbf{Metabuli}} & \multicolumn{3}{c}{\textbf{MMseqs2}} & \multicolumn{3}{c}{\textbf{Kraken2}} \\
\cmidrule(lr){3-5} \cmidrule(lr){6-8} \cmidrule(lr){9-11}
& & Correct & Wrong & Unlbl. & Correct & Wrong & Unlbl. & Correct & Wrong & Unlbl. \\
\midrule

% ==================== 第 1 个数据集 ====================
\multirow{5}{*}{\begin{tabular}[c]{@{}l@{}}\textbf{\textit{Gastrointestinal}}\\ (N=81,602)\end{tabular}} 
& Base                     & 52,060 & 4,903 & 4,683 & 52,060 & 2,774 & 26,768 & 67,714 & 9,398 & 4,490 \\
& Taxometer                & 74,885 & 4,671 & 2,046  & 74,354 & 5,033 & 2,215  & 66,355 & 8,819 & 6,428 \\
& Taxometer\_G             & 74,978 & 3,693 & 2,931  & 74,231 & 3,844 & 3,527  & 65,195 & 7,389 & 9,018 \\
& TaxDistill (100M)        & 75,452 & 4,379 & 1,771  & 75,952 & 3,902 & 1,748  & 66,592 & 8,497 & 6,513 \\
& TaxDistill (500M)        & 75,481 & 4,442 & 1,679  & 75,969 & 3,858 & 1,775  & 66,651 & 8,342 & 6,609 \\
\midrule

% ==================== 第 2 个数据集 ====================
\multirow{5}{*}{\begin{tabular}[c]{@{}l@{}}\textbf{\textit{Airways}}\\ (N=187,685)\end{tabular}} 
& Base                     & 164,799 & 11,966 & 10,920 & 117,170 & 5,863 & 64,652 & 158,516 & 15,499 & 13,670 \\
& Taxometer                & 168,511 & 12,325 & 6,849 & 164,925 & 11,964 & 10,796 & 153,008 & 14,377 & 20,300 \\
& Taxometer\_G             & 165,574 & 11,387 & 10,724 & 161,426 & 11,522 & 14,737 & 150,212 & 13,882 & 23,591 \\
& TaxDistill (100M)        & 167,438 & 8,330  & 11,917 & 166,631 & 7,995 & 13,059 & 148,447 & 13,265 & 25,973 \\
& TaxDistill (500M)        & 167,508 & 8,368  & 11,809 & 166,804 & 7,927 & 12,954 & 148,974 & 13,183 & 25,528 \\
\midrule

% ==================== 第 3 个数据集 ====================
\multirow{5}{*}{\begin{tabular}[c]{@{}l@{}}\textbf{\textit{Skin}}\\ (N=173,927)\end{tabular}} 
& Base                     & 152,995 & 11,016 & 9,916  & 107,964 & 5,360  & 60,603 & 150,894 & 14,601 & 8,432 \\
& Taxometer                & 153,276 & 10,404 & 10,247  & 152,848 & 10,839 & 10,240 & 146,548 & 13,728 & 13,651 \\
& Taxometer\_G             & 152,097 & 9,465  & 12,365 & 149,149 & 10,724 & 14,054 & 143,225 & 13,329 & 17,373 \\
& TaxDistill (100M)        & 153,725 & 6,559  & 13,643  & 153,581 & 7,195  & 13,151 & 142,949 & 12,589 & 18,389 \\
& TaxDistill (500M)        & 153,650 & 6,417  & 13,860  & 153,683 & 7,462  & 12,782 & 142,766 & 12,680 & 18,481 \\
\midrule

% ==================== 第 4 个数据集 ====================
\multirow{5}{*}{\begin{tabular}[c]{@{}l@{}}\textbf{\textit{Oral}}\\ (N=201,563)\end{tabular}} 
& Base                     & 175,800 & 13,711 & 12,052 & 122,913 & 7,096  & 71,554 & 176,861 & 18,153 & 6,549 \\
& Taxometer                & 180,492 & 13,321 & 7,750  & 178,898 & 13,381 & 9,284  & 174,300 & 16,689 & 10,574 \\
& Taxometer\_G             & 177,836 & 12,402 & 11,325 & 175,598 & 12,855 & 13,110 & 172,035 & 16,189 & 13,339 \\
& TaxDistill (100M)        & 180,853 & 9,606  & 11,104 & 180,730 & 9,562  & 11,271 & 173,794 & 15,623 & 12,146 \\
& TaxDistill (500M)        & 181,163 & 9,539  & 10,861 & 180,956 & 9,691  & 10,916 & 173,665 & 15,491 & 12,407 \\
\midrule

% ==================== 第 5 个数据集 ====================
\multirow{5}{*}{\begin{tabular}[c]{@{}l@{}}\textbf{\textit{Urogenital}}\\ (N=57,529)\end{tabular}} 
& Base                     & 50,065 & 3,826 & 3,638 & 35,626 & 2,030 & 19,873 & 50,551 & 6,208 & 770 \\
& Taxometer                & 50,920 & 4,386 & 2,223 & 49,875 & 4,760 & 2,894  & 49,888 & 5,967 & 1,674 \\
& Taxometer\_G             & 50,394 & 4,314 & 2,821 & 48,820 & 4,743 & 3,966  & 49,873 & 5,875 & 1,781 \\
& TaxDistill (100M)        & 51,288 & 3,939 & 2,302 & 50,413 & 4,844 & 2,272  & 50,410 & 5,894 & 1,225 \\
& TaxDistill (500M)        & 51,357 & 3,944 & 2,228 & 50,404 & 4,804 & 2,321  & 50,396 & 5,898 & 1,235 \\
\midrule

% ==================== 第 6 个数据集 ====================
\multirow{5}{*}{\begin{tabular}[c]{@{}l@{}}\textbf{\textit{Rhizosphere}}\\ (N=28,318)\end{tabular}} 
& Base                     & 22,114 & 5,881 & 323   & 16,603 & 2,957 & 8,758  & 16,125 & 8,237 & 3,956 \\
& Taxometer                & 22,125 & 5,878 & 315    & 21,703 & 6,204 & 411    & 16,025 & 5,225 & 7,068 \\
& Taxometer\_G             & 22,076 & 5,841 & 401  & 21,689 & 6,051 & 578    & 16,008 & 4,958 & 7,352 \\
& TaxDistill (100M)        & 21,966 & 5,191 & 1,161 & 21,828 & 5,194 & 1,296  & 15,989 & 3,120 & 9,209 \\
& TaxDistill (500M)        & 21,962 & 5,155 & 1,201 & 21,823 & 5,235 & 1,260  & 15,986 & 3,217 & 9,115 \\
\midrule

% ==================== 第 7 个数据集 ====================
\multirow{5}{*}{\begin{tabular}[c]{@{}l@{}}\textbf{\textit{Marine}}\\ (N=216,508)\end{tabular}} 
& Base                     & 178,531 & 30,330 & 7,647  & 141,296 & 22,056 & 53,156 & 153,800 & 42,315 & 20,393 \\
& Taxometer                & 180,839 & 29,034 & 6,635  & 180,320 & 28,839 & 7,349  & 151,720 & 33,883 & 30,905 \\
& Taxometer\_G             & 179,695 & 27,644 & 9,169  & 178,226 & 28,104 & 10,178 & 151,825 & 32,200 & 32,483 \\
& TaxDistill (100M)        & 180,784 & 20,510 & 15,214 & 182,611 & 23,832 & 10,065 & 151,521 & 27,763 & 37,224 \\
& TaxDistill (500M)        & 180,863 & 20,325 & 15,320 & 183,023 & 23,723 & 9,762  & 151,946 & 27,588 & 36,974 \\
\bottomrule
\end{tabular}
}
\end{table*}

\end{document}